\documentclass[10pt,twocolumn,letterpaper]{article}

\usepackage[pagenumbers]{cvpr} 

\usepackage{blindtext}
\usepackage{graphicx}
\usepackage{amsmath}
\usepackage{amssymb}
\usepackage{booktabs}
\usepackage{multirow}
\usepackage{float}
\usepackage{bm}
\usepackage{esint}
\usepackage{float}
\usepackage{graphicx}
\usepackage{colortbl}
\usepackage{xcolor}
\usepackage{array}
\definecolor{mygray}{RGB}{220,220,220}

\usepackage[pagebackref,breaklinks,colorlinks]{hyperref}

\usepackage[capitalize]{cleveref}
\crefname{section}{Sec.}{Secs.}
\Crefname{section}{Section}{Sections}
\Crefname{table}{Table}{Tables}
\crefname{table}{Tab.}{Tabs.}

\def\cvprPaperID{6384} 
\def\confName{CVPR}
\def\confYear{2023}

\begin{document}



\title{A Unified Mutual Supervision Framework for Referring Expression Segmentation and Generation
}

\author{\textbf{Shijia Huang$^{1}$}\thanks{Internship work at IDEA. 
}, \textbf{Feng Li$^{2,3*}$}, \textbf{Hao Zhang$^{2,3*}$}, \textbf{Shilong Liu$^{2,4*}$}, \textbf{Lei Zhang$^{2}$}\thanks{Corresponding author.}, \textbf{Liwei Wang$^{1}$} \and
$^{1}$The Chinese University of Hong Kong. 
$^{2}$International Digital Economy Academy (IDEA). \\
$^{3}$The Hong Kong University of Science and Technology. 
$^{4}$Tsinghua University. \\
 {\tt\small \{sjhuang,lwwang\}@cse.cuhk.edu.hk}, {\tt\small \{fliay,hzhangcx\}@connect.ust.hk}, \\
 {\tt\small \{liusl20\}@mails.tsinghua.edu.cn},
 {\tt\small \{leizhang\}@idea.edu.cn}
}

\maketitle

\begin{abstract}

Reference Expression Segmentation (RES) and Reference Expression Generation (REG) are mutually inverse tasks that can be naturally jointly trained. Though recent work has explored such joint training, the mechanism of how RES and REG can benefit each other is still unclear. In this paper, we propose a unified mutual supervision framework that enables two tasks to improve each other. Our mutual supervision contains two directions. On the one hand, Disambiguation Supervision leverages the expression unambiguity measurement provided by RES to enhance the language generation of REG. On the other hand, Generation Supervision uses expressions automatically generated by REG to scale up the training of RES. Such unified mutual supervision effectively improves two tasks by solving their bottleneck problems. Extensive experiments show that our approach significantly outperforms all existing methods on REG and RES tasks under the same setting, and detailed ablation studies demonstrate the effectiveness of all components in our framework.
\end{abstract}

\section{Introduction}
\label{sec:intro}
Referring expression and its related tasks~\cite{yu2016modeling,mao2016generation,hu2016segmentation,yu2017joint,luo2020multi,ding2021vision,tanaka2019generating} have attracted increasing interest of the vision-language community in recent years.
Among them, Referring Expression Segmentation (RES)~\cite{hu2016segmentation,luo2020multi,ding2021vision} aims to find a target object in an image given a query expression and outputs the segmentation mask. Conversely, Referring Expression Generation (REG)~\cite{mao2016generation,yu2016modeling,tanaka2019generating} seeks to generate a natural language expression for a specified object. These tasks are fundamental as building blocks for vision-language techniques and are crucial for many applications~\cite{lopes2000human,doshi2008spoken,xia2021tedigan,li2020manigan}.


\begin{figure}[t]
  \centering
  \includegraphics[width=0.45\textwidth]{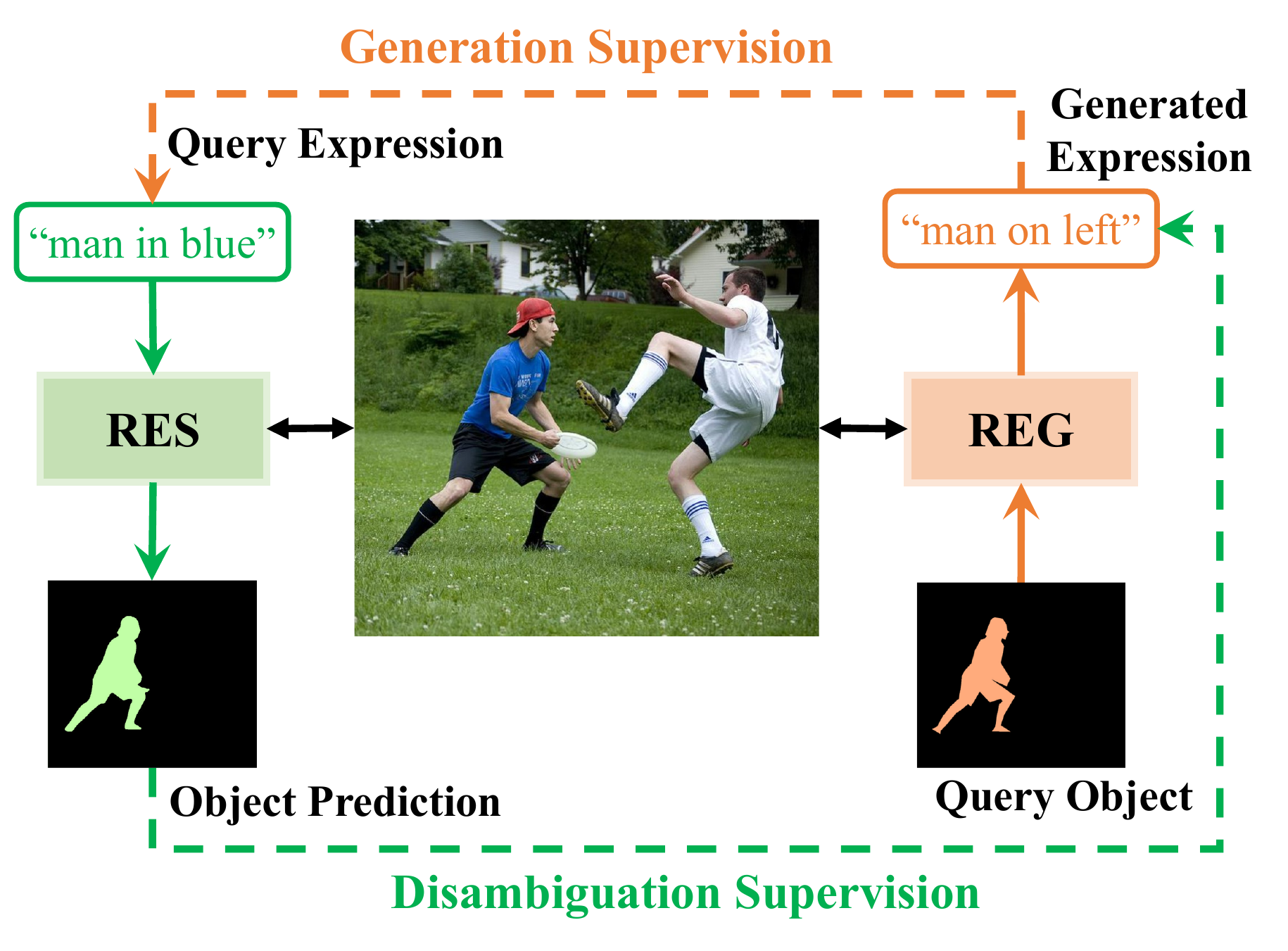}
  \vspace{-10pt}
  \caption{Around the referring expression, the derived Referring Expression Segmentation (RES) and Referring Expression Generation (REG) are inverse tasks. We propose a novel mutual supervision to improve each other.}
    \vspace{-15pt}
  \label{fig:intro}
\end{figure}

The RES task relies on paired annotations, \textit{i.e.}, pixel-level masks matched with language descriptions. Existing datasets like RefCOCO~\cite{yu2016modeling}, RefCOCO+~\cite{yu2016modeling}, and RefCOCOg~\cite{mao2016generation}, augment language descriptions to MS-COCO instance segmentation dataset~\cite{lin2014microsoft}. However, due to the costly annotation process, only $71$k out of $886$k instances are finally labeled with language descriptions. Describing the specific instance in the image, REG has a higher requirement on the generated language, which should refer back to the target object without any ambiguity. Nevertheless, existing learning objectives for this task~\cite{mao2016generation,yu2016modeling,liu2017referring} usually neglect this characteristic without putting adequate effort into enhancing generation results.

Recent work~\cite{mao2016generation,yu2016modeling,tanaka2019generating,sun2022proposal} has explored learning RES and REG tasks together since they use the same paired data. However, most of them focus on studying different architectures while neglecting to investigate the inherent problem: \textit{How can the RES and REG benefit each other in a joint learning framework?} In this paper, we will try to answer this question. 
As shown in Fig~\ref{fig:intro}, on the one hand, given a generated expression from the REG model, the unambiguity of the expression can be reflected when a RES model can accurately find the referred object by taking this expression as the query. On the other hand, the REG model can automatically generate large amounts of expressions to increase the training scale of the RES model, which helps the RES task get rid of extensive human annotations. Therefore, effectively using these characteristics can alleviate the inherent problems of RES and REG tasks and, thus, motivates us to explore a mutual supervision paradigm to enhance two tasks jointly.



To this end, we propose a transformer-based unified mutual supervision framework, including a proposal extractor, an indicated generation head for REG, and a proposal selection head for RES. Among them, a novel indicator module is also proposed to specify objects and instruct the expression generation process. Our framework learns under mutual supervision from two tasks, \textit{i.e.}, Disambiguation Supervision (RES $\rightarrow$ REG) and Generation Supervision (REG $\rightarrow$ RES).
In Disambiguation Supervision (RES $\rightarrow$ REG), the proposal selection head measures the unambiguity of the generated expression by comparing its matching score to the target object with the context. We apply reinforcement learning with this unambiguity reward to improve the REG task.
While in the Generation Supervision (REG $\rightarrow$ RES), we leverage the indicated generation head to automatically generate expressions for the instance segmentation data, thus improving the RES task by scaling up its dataset. The proposed mutual supervision can effectively disambiguate generated expressions of REG and alleviate the lack of data in RES.



We conduct detailed experiments on three popular datasets, \textit{i.e.}, RefCOCO~\cite{yu2016modeling}, RefCOCO+~\cite{yu2016modeling}, and RefCOCOg~\cite{mao2016generation}. 
Our approach significantly outperforms all existing methods on both REG and RES tasks under the same settings. For the REG task, expressions generated by our model present superior performance under both automatic metrics and human evaluation. 
Especially for the hardest split RefCOCO+ testB, we improve the CIDEr metric from the previous best of 0.860 to 0.927.
For the RES task, we surpass previous methods by a large margin under both oIoU and mIoU metrics, and even better than RefTR~\cite{li2021referring} with large Visual Genome~\cite{krishna2017visual} pretraining. 
Extensive ablation studies demonstrate the effectiveness of our architecture as well as the proposed mutual supervision.

\section{Related Work}

\noindent \textbf{Referring Expression Segmentation (RES).}
Referring Expression Comprehension (REC)~\cite{yang2019fast,deng2021transvg}, and visual grounding~\cite{plummer2015flickr30k,deng2018visual,wang2016learning,wang2021improving,huang2022multi} aims to predict bounding boxes of referred objects based on a language query. RES~\cite{hu2016segmentation,hui2020linguistic,margffoy2018dynamic} is a similar task but outputs a pixel-wise segmentation mask, which is usually more challenging. Recent approaches~\cite{liu2017recurrent,chen2019see,ye2019cross,li2018referring,luo2020multi,luo2020cascade,jing2021locate,ding2021vision,li2021referring,huang2020referring,hu2020bi} follow a bottom-up pipeline that fuses multi-modal features pixel-by-pixel and produces segmentation masks directly. Among them, some works~\cite{luo2020multi,luo2020cascade,jing2021locate,yang2021bottom,feng2021encoder} focus on modeling the multi-modal interaction, and others~\cite{ding2021vision,li2021referring,yang2021lavt,wang2021cris} explore more powerful Transformer-based architectures, as well as using stronger backbones like Swin-Transformer~\cite{liu2021swin}. In parallel, 
another line of work adopts a top-down pipeline~\cite{yu2018mattnet,chen2019referring,liu2019learning}, which first extracts all instance proposals from the image and then selects the best matching proposal. In this work, we propose a novel Transformer-based RES method in a top-down manner, employing a powerful DETR-like segmentation network as a proposal extractor and a Transformer-Decoder architecture for proposal selection.

\noindent \textbf{Referring Expression Generation (REG).}
REG~\cite{mao2016generation,niu2019variational,yu2017joint,liu2020attribute,liu2017referring} is the inverse task of RES and REC, aiming to generate unambiguous language descriptions for the mentioned objects. To reduce the ambiguity, some approaches~\cite{niu2019variational,yu2017joint} calculate the difference of object features as additional information, while others~\cite{liu2020attribute,liu2017referring} explicitly learn to predict object attributes. 
Instead, we propose a novel indicator module to flexibly instruct the role of each object in the expression generation, making the generation process more accurate and controllable.
While previous works~\cite{mao2016generation,yu2016modeling,yu2017joint,liu2017referring,tanaka2019generating,sun2022proposal} jointly train the REG and RES or REC tasks, the performance gains usually come from sophisticated 
network architecture design, while we focus on investigating the inherent problem of how RES and REG tasks can benefit each other. In this paper, we propose a novel mutual supervision framework to achieve this goal.

\begin{figure*}[htbp]
  \centering
  \includegraphics[width=0.78\textwidth]{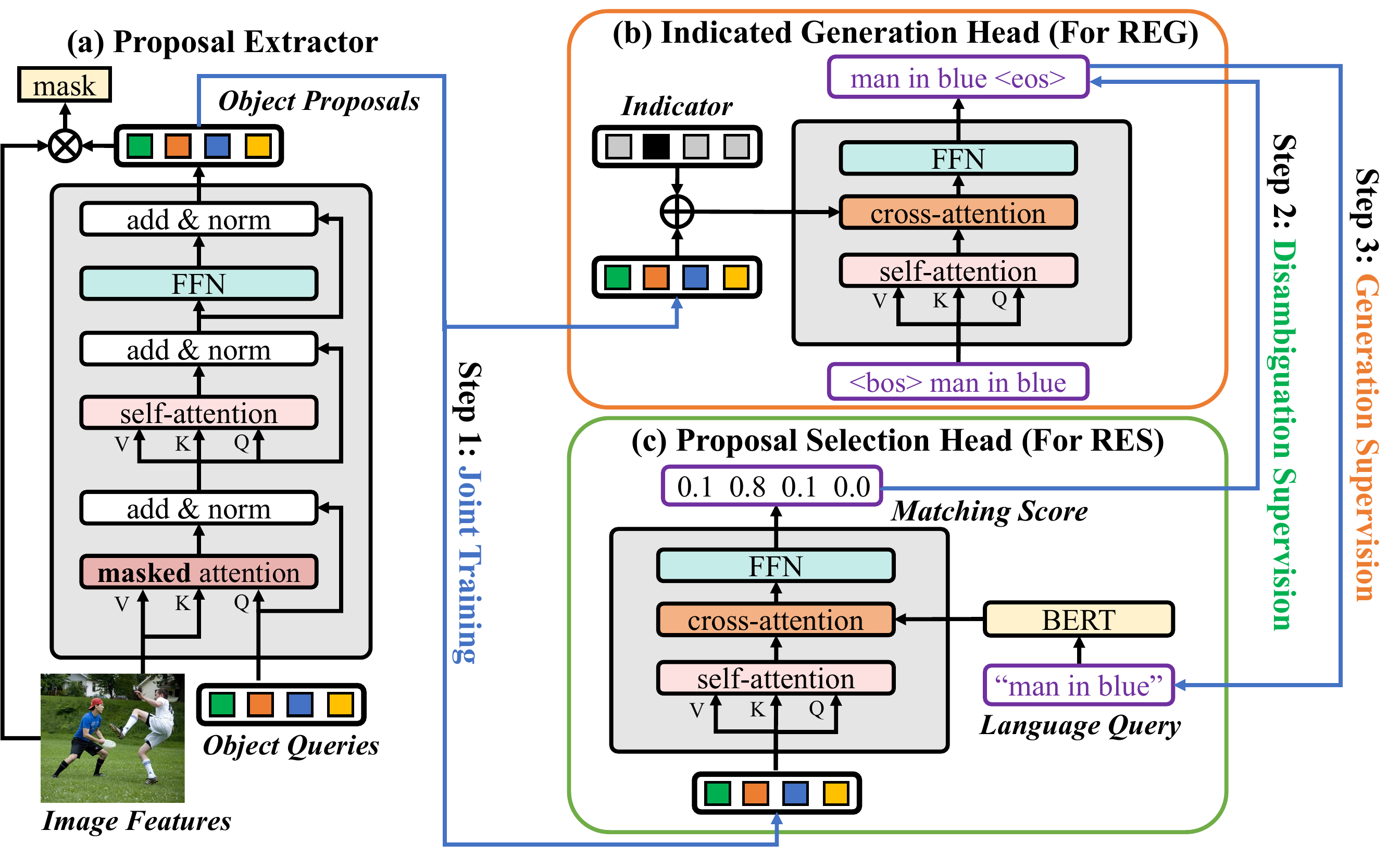}
  \vspace{-5pt}
  \caption{The framework of our proposed network. Based on the DETR-like segmentation network for proposal extraction, we propose an indicated generation head for REG and a proposal selection head for RES. `add \& norm' blocks are omitted in (b) (c).}
  \vspace{-10pt}
  \label{fig:framework}
\end{figure*}

\noindent \textbf{Scaling Up Referring Expressions.}
Previous work pays efforts to scale up referring expressions from different aspects. PhraseCut~\cite{wu2020phrasecut} proposes a template-based dataset for language-based image segmentation based on categories, attributes, and relationships from Visual Genome~\cite{krishna2017visual}. While Pseudo-Q~\cite{jiang2022pseudo} generates pseudo expressions based on manually designed prompts and off-the-shelf detectors.
Besides data generation, RefTR~\cite{li2021referring} performs large-scale pretraining on region descriptions in Visual Genome to improve both REC and RES. While MDETR~\cite{kamath2021mdetr} and VL-T5~\cite{cho2021unifying} conduct pretraining by merging multiple vision-language tasks.
A concurrent work UniRef~\cite{zheng2022towards} also adopts similar pretraining with additional large-scale data~\cite{krishna2017visual} into the REC and REG joint training. Since these pretraining works use a dozen times more human-labeled data and more extensive training resources, they cannot be directly comparable with the settings~\cite{yu2018mattnet,luo2020multi,yu2017joint,liu2017referring} that we follow in this paper. Instead of leveraging external human-annotated data, we scale up the RES training by automatically generating expressions from the interleaved REG task in our mutual supervision framework.

\noindent \textbf{Improving Vision-based Text Generation.}
Since text generation is a discrete sampling process, optimizing the network directly through the generated sentences is hard. 
To tackle this problem, SCST~\cite{rennie2017self} employs reinforcement learning~\cite{sutton2018reinforcement} with the CIDEr score as a reward and optimizes the network according to policy gradient~\cite{williams1992simple}. Other works~\cite{xu2017semi,dognin2019improved} employ Gumbel-Softmax approximation to preserve gradients of the sampling and apply adversarial/reconstruction loss for optimization.
Previous work JointSLR~\cite{yu2017joint} and following works~\cite{tanaka2019generating,kim2020conan} improve REG through reinforcement learning, in which they additionally train a simple network to score the generated expression. Different from them, we leverage the jointly trained RES network to provide disambiguation supervision. Our unambiguity reward compares matching scores of the expression with the target object and with those in the context, reflecting the expression's unambiguity comprehensively.


\section{Approach}

\subsection{Task Description}
\label{sec:task_define}

\noindent \textbf{Referring Expression Generation (REG).} REG requires to generate an expression $Y$ for a given target object $O_{t}$ in the image. REG has higher requirements on the expression unambiguity, \textit{i.e.}, $Y$ must distinguish $O_{t}$ from other objects.

\noindent \textbf{Referring Expression Segmentation (RES).} Given an image and a referring expression $Y$, the goal of RES is to find out the target object $O_{t}$ that matches the description of $Y$ and outputs the corresponding segmentation mask $M_{t}$.

\subsection{Model Architecture}
\label{sec:1}

Our model follows a top-down pipeline~\cite{yu2018mattnet}. As shown in Fig.~\ref{fig:framework}, given the image features from the visual backbone, the proposal extractor first detects objects in the image as object proposals. Then we design an indicated generation head for REG and a proposal selection head for RES. We will go over these modules one by one.

\noindent \textbf{Visual Backbone.}
We adopt ResNet-101~\cite{he2016deep} as the visual backbone.
Given an image in the shape of $H \times W$, the visual features from multiple stages of ResNet are denoted as $F_{v_i} \in \mathbb{R}^{C_{i} \times H_{i} \times W_{i}}$, where $H_{i}$ and $W_{i}$ are the downsampled width and height of $H$ and $W$, with a rate of $\{4,8,16,32\}$. 

\noindent \textbf{Proposal Extractor.} 
We adopt Mask2Former~\cite{cheng2021masked} as the proposal extractor, which is a DETR-like~\cite{carion2020end} segmentation model. In Mask2Former, a transformer encoder is first applied on the multi-stage visual features $F_{v_{i}}$.
Fig.~\ref{fig:framework} (a) shows the decoder part of Mask2Former, the inputs are the encoded image features and $N$ learnable object queries, $N$ is usually set to $100$.
Then in each decoder layer, the object queries extract visual features for different objects using the attention mechanism. 
The output of the proposal extractor are $N$ object proposals $\mathcal{O}=[O_{1},...,O_{N}]$, where $O_{i} \in \mathbb{R}^{C}$ contains rich visual features of the $i$-th object.
The binary segmentation masks $[M_{1},..., M_{N}]$ are predicted by the dot product between $\mathcal{O}$ and the high-resolution visual feature $F_\mathrm{pixel}$ from a pixel decoder. The training of the proposal extractor follows Mask2Former.

\begin{figure}[t]
  \centering
  \includegraphics[width=0.43\textwidth]{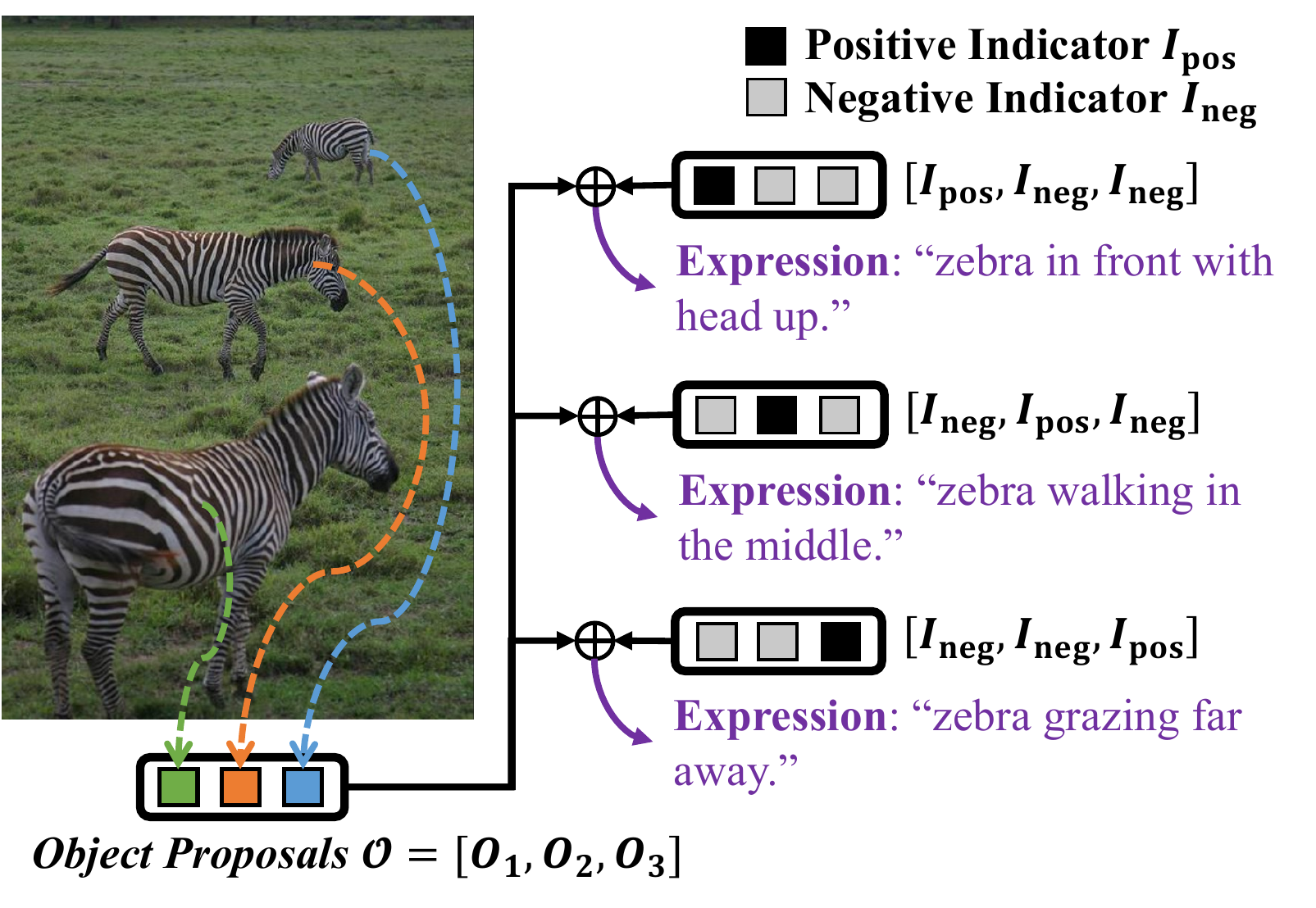}
  \vspace{-5pt}
  \caption{Illustration of the proposed indicator module.}
    \vspace{-10pt}
  \label{fig:indicator}
\end{figure}

\noindent \textbf{Indicated Generation Head.} 
Fig.~\ref{fig:framework} (b) shows the proposed indicated generation head for REG. We adopt the regular transformer decoder architecture, which queries object features via cross-attention and predicts expressions in an autoregressive manner.
Following previous works ~\cite{sun2022proposal,mao2016generation,liu2017referring,liu2020attribute} that rely on only the target object to generate referring expressions, given all object proposal $\mathcal{O}$ and to describe $O_{k}$, we can simply deprecate $O_{i \neq k}$.
However, we claim that fully utilizing these negative samples $O_{i \neq k}$ is helpful to generate unambiguous expressions, and propose a novel \textbf{Indicator} module, given as:
\begin{equation}
   \mathcal{I}(k) = [\underbrace{I_\mathrm{neg},\dots,I_\mathrm{neg}}_{k-1},I_\mathrm{pos},\underbrace{I_\mathrm{neg},\dots,I_\mathrm{neg}}_{N-k}],
\end{equation}
\noindent where the positive indicator $I_\mathrm{pos} \in \mathbb{R}^{C}$ and the negative indicator $I_\mathrm{neg} \in \mathbb{R}^{C}$ are independent learnable embeddings, which instruct the role (positive/negative) of each object proposal, \textit{i.e.}, the generated expression should correspond to the positive object while distinguishing from the negative objects.
Hence the expression generation for object $k$ is based on the indicated objects $\mathcal{O}+\mathcal{I}(k)$, in which we assign $I_{pos}$ to $O_{k}$ and assign $I_{neg}$ to all others.
The assignment is achieved by simple feature addition, and Fig.~\ref{fig:indicator} shows the example of different indicator assignments.
The training of the indicated generation head follows teacher-forcing learning~\cite{bengio2015scheduled}.
Given a ground-truth expression $Y^{*}$ of object $k$, we minimize the following cross-entropy loss:
\begin{equation}
    \mathcal{L}_\mathrm{gen} = -\sum_{l}^{L} \log p(Y^{*}_{l}|Y^{*}_{<l},\mathcal{O}+\mathcal{I}(k)),
    \label{eq:l_reg}
\end{equation}
\noindent where $L$ is the length of $Y^{*}$, and $Y^{*}_{<l}$ is a prefix of $Y^{*}$. 
At inference time, we map the instance masks to object proposals by bipartite matching and predict the expression word by word with a $<$bos$>$ token as the beginning.

\noindent \textbf{Proposal Selection Head.} We model the RES task as a proposal selection problem, i.e., calculating the matching scores $\mathcal{S}=[S_{1},..., S_{N}]$ of all proposals $\mathcal{O}$ and select the one with the highest score.
As shown in Fig.~\ref{fig:framework} (c), the proposal selection head adopts a regular transformer decoder architecture. Given the referring expression $Y$, we adopt $\text{BERT}_\mathrm{base}$~\cite{devlin2018bert} to obtain the text feature $F_{Y} \in \mathbb{R}^{L \times C}$, where $L$ is the length of $Y$. Then the matching score $\mathcal{S}$ of the expression $Y$ is given as:
\begin{equation}
    \mathcal{S}(Y) =  \mathrm{MLP}(\mathrm{Decoder}(\mathcal{O},F_{Y})),
    \label{eq:matching_score}
\end{equation}
\noindent in which the object proposals interact with each other in the self-attention layer and query $F_{Y}$ in the cross-attention layer. We adopt a cross-entropy loss to supervise the training of $\mathcal{S}$. At inference time, we output the mask $M_{k}$ of the object proposal $O_{k}$ with the largest matching score $S_{k}$.

\subsection{Mutual Supervision}

\subsubsection{Disambiguation Supervision (RES $\rightarrow$ REG)}

As the task defined in Sec.~\ref{sec:task_define}, unambiguity is important for REG. For example, given an image of many zebras, the expression ``zebra with many stripes'' is ambiguous for one zebra because we cannot tell which zebra it is describing.
However, the typical training objective in Eq.(\ref{eq:l_reg}) does not guarantee the expression unambiguity at inference time.
Inspired by self-critical sequence training~\cite{rennie2017self} in image captioning task, we leverage reinforcement learning~\cite{sutton2018reinforcement} to improve the generation ability but focus on unambiguity.

Specifically, given the indicated generation head with parameter $\theta$, an expression $Y$ for object $k$ is generated under the distribution $p_{\theta}(Y,k) = \prod_{l}^{L} p_{\theta}(Y_{l}|Y_{<l},\mathcal{O}+\mathcal{I}(k))$.
Our goal is to maximize the expected reward $r(Y,k)$, 
according to policy gradient~\cite{williams1992simple}, the gradient of $\theta$ is given as:
\begin{equation}
    \nabla_{\theta} \mathcal{L}_\mathrm{R} = -\mathbb{E}_{Y \sim p_{\theta}}[r(Y,k) \nabla_{\theta} \log p_{\theta}(Y,k)],
    \label{eq:reinforce}
\end{equation}
\noindent The matching score $\mathcal{S}$ provided by proposal selection head in Eq.(\ref{eq:matching_score}) can be strong supervision, \textit{i.e}., the expression of object $k$ should have the greatest matching score with $k$ across all objects. Hence we design an unambiguity reward by considering matching scores across all objects, given as:
\begin{equation}
    r(Y,k) = \frac{\exp(S_{k}(Y))}{\sum_{i=1}^{N}\exp(S_{i}(Y))}.
\end{equation}
\noindent The supervision from the proposal selection head is noisy. 
First, the matching score is sometimes overconfident, resulting in rewards of multiple expressions close to $1$, and hard to distinguish which expression is better. We thus introduce a factor $\tau$ to smooth $\mathcal{S}$. 
Second, we find the unambiguity reward hurts the performance under automatic metrics (e.g, CIDEr~\cite{vedantam2015cider}), we thus introduce the CIDEr score in the reward to balance the learning. The final reward is given as:
\begin{equation}
    r(Y,k) = \alpha \cdot \frac{\exp(S_{k}(Y)/\tau)}{\sum_{i=1}^{N}\exp(S_{i}(Y)/\tau)} + \mathrm{CIDEr}(Y,Y^{*}).
    \label{eq:reward}
\end{equation}
\noindent where $Y^{*}$ is ground-truth expressions for object $k$, $\alpha$ balances two rewards. In practice, we sample $q_{1}$ expressions by beam search~\cite{freitag2017beam} to approximate the gradient in Eq.(\ref{eq:reinforce}).

\subsubsection{Generation Supervision (REG $\rightarrow$ RES)}

Data annotation for segmentation masks and referring expressions are both labor-intensive, which causes data shortage in RES.
We notice that there are still $87$k of COCO~\cite{lin2014microsoft} training images with instance segmentation annotations but no referring expression annotations, and apply the indicated generation head to generate expressions on these data as the generation supervision.

Specifically, we generated $q_{2}$ expressions for each unannotated instance using beam search~\cite{freitag2017beam}, denote as $D_\mathrm{gen}$. Then we can directly train the RES model on a larger dataset by combining $D_\mathrm{gen}$ with human-annotated data $D_\mathrm{real}$. Since $D_\mathrm{gen}$ is noisy, we implement the following two strategies to get better performance: 
(1) \textit{Data Filtering}: We observe that annotated instances in existing RES datasets are usually large objects, and the description accuracy of the REG model for small objects is much lower than that of large objects. Hence we design area-based rules to filter small objects. 
(2) \textit{Data Reweighting}: Even for large objects, there is still noise in the generated expressions. We down-weight the generated expressions in the loss function to further suppress the noise.

\subsection{Training Scheme}
We train our model in three steps. In step 1, we jointly train the entire network for both REG and RES tasks end-to-end. In step 2, we utilize disambiguation supervision to train the indicated generation head while freezing other parts. In step 3, we generate expressions and utilize generation supervision to retrain the network. We report REG performance at step 2 and RES performance at step 3.

\section{Experiments}

\subsection{Experimental Setup}

\noindent \textbf{Datasets.} We conduct extensive experiments on three benchmark datasets for both REG and RES, including RefCOCO~\cite{yu2016modeling}, RefCOCO+~\cite{yu2016modeling}, and RefCOCOg~\cite{mao2016generation}, which are all collected from MS-COCO~\cite{lin2014microsoft}. A detailed introduction of the datasets is in the Appendix.

\noindent \textbf{Evaluation Metrics for REG.}
Following previous works~\cite{sun2022proposal,tanaka2019generating,yu2017joint,liu2020attribute}, we adopt common automatic metrics Meteor~\cite{banerjee2005meteor} and CIDEr~\cite{vedantam2015cider} for generated expressions evaluation. And following~\cite{yu2017joint,tanaka2019generating,sun2022proposal} to compute CIDEr robustly, we merge the expressions from RefCOCO and RefCOCO+ to expand their ground-truth expression set.
To evaluate the expression unambiguity, we conduct the human evaluation following previous works~\cite{mao2016generation,liu2017referring,liu2020attribute,sun2022proposal}. 


\noindent \textbf{Evaluation Metrics for RES.}
We adopt the common metrics of overall intersection-over-union (oIoU) and mean intersection-over-union (mIoU) following previous works~\cite{luo2020multi,jing2021locate,li2021referring}. The oIoU divides the total intersection area by the total union area of all test samples. While mIoU takes the average of the IoU of all test samples. We also adopt the precision with different thresholds (Pr@X), which measures the percentage of predictions whose IoU$>$X.

\begin{figure}[htbp]
  \centering
  \includegraphics[width=0.45\textwidth]{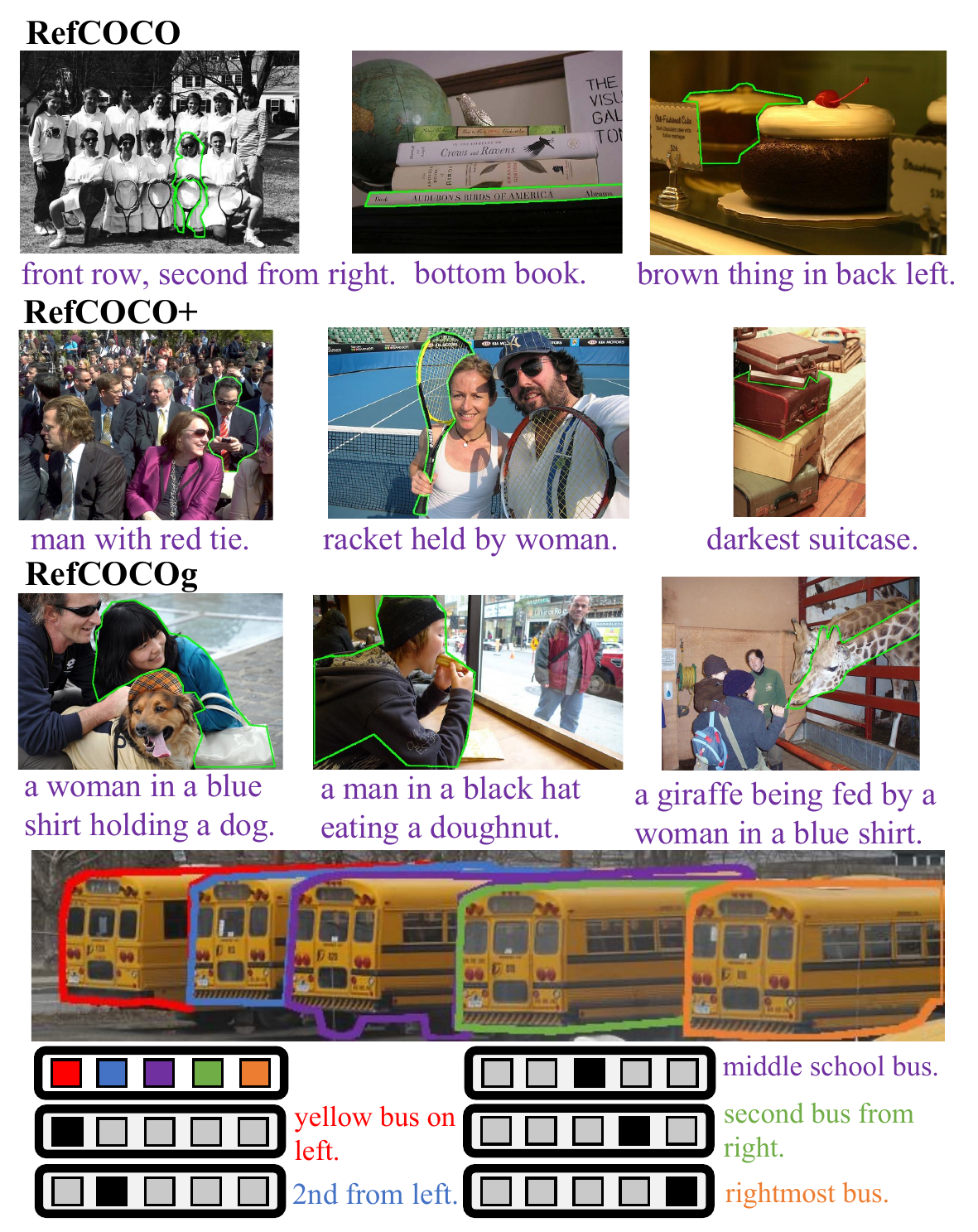}
  \vspace{-5pt}
  \caption{Qualitative Results for our REG model.}
  \vspace{-15pt}
  \label{fig:vis}
\end{figure}

\begin{table*}[htbp]
\centering
\footnotesize
\resizebox{16.0cm}{!}{
\begin{tabular}{l|c|c|c|c|c|c|c|c|c|c|c}
\hline \toprule
 \multirow{3}{*}{Methods} & \multirow{3}{*}{Input} & \multicolumn{4}{c|}{RefCOCO} & \multicolumn{4}{c|}{RefCOCO+} & \multicolumn{2}{c}{RefCOCOg} \\ \cline{3-12}
 & & \multicolumn{2}{c|}{testA} & \multicolumn{2}{c|}{testB} & \multicolumn{2}{c|}{testA} & \multicolumn{2}{c|}{testB} &  \multicolumn{2}{c}{val-g} \\ \cline{3-12}
 & & Meteor & CIDEr & Meteor & CIDEr & Meteor & CIDEr & Meteor & CIDEr & Meteor & CIDEr \\
 \hline \toprule
 MMI~\cite{mao2016generation} & Box & 0.243	& 0.615 & 0.300 & 1.227 & 0.199 & 0.462 & 0.189 & 0.679 & 0.149 & 0.585\\
 MCRE~\cite{yu2016modeling} & Box & 0.260 & 0.679 & 0.319 & 1.276 & 0.202 & 0.475 & 0.196 & 0.683 & 0.147 & 0.573 \\
 Attr~\cite{liu2017referring} & Box & 0.274 & 0.710 & 0.313 & 1.257 & 0.219 & 0.512 & 0.203 & 0.704 & 0.157 & 0.639 \\
 VC~\cite{niu2019variational} & Box & 0.188 & 0.707 & 0.245 & 1.356 & 0.142 & 0.518 & 0.146 & 0.731 & 0.139 & 0.625 \\
 Attention~\cite{liu2020attribute}  & Box & 0.312 & 0.802 & 0.332 & 1.301 & 0.236 & 0.585 & 0.206 & 0.692 & 0.163 & 0.645  \\
 Joint SLR~\cite{yu2017joint} & Box & 0.296 & 0.775 & 0.340 & 1.320 & 0.213 & 0.520 & 0.215 & 0.735 & 0.159 & 0.662  \\
 GERE~\cite{tanaka2019generating} & Box & 0.310 & 0.859 & 0.342 & 1.375 & 0.241 & 0.663 & 0.225 & 0.812 & 0.164 & 0.763  \\
 PFOS~\cite{sun2022proposal} & Box & 0.304 & 0.872 & 0.324 & 1.315 & 0.250 & 0.719 & 0.211 & 0.767 & 0.157 & 0.754 \\
 CoNAN~\cite{kim2020conan} & Box & 0.328 & 0.912 & 0.351 & 1.422 & \underline{0.281} & 0.750 & 0.243 & 0.860 & \underline{0.180} &  \underline{0.905} \\ \hline
 Ours & Box & \underline{0.331} & \underline{0.941} & \underline{0.365} & \underline{1.472} & 0.258 & \underline{0.764 } & \underline{0.243} & \underline{0.886} & 0.171 & 0.833  \\ 
 Ours & Mask & \textbf{0.347} & \textbf{1.023} & \textbf{0.377} & \textbf{1.527} & \textbf{0.286} & \textbf{0.820} & \textbf{0.248} & \textbf{0.927} & \textbf{0.184} & \textbf{0.926}  \\
 \bottomrule \hline
\end{tabular}
}
\vspace{-5pt}
\caption{Comparison with state-of-the-art methods for Referring Expression Generation (REG), under Meteor and CIDEr metric.}
\vspace{-15pt}
\label{tab:reg_main_tab}
\end{table*}

\noindent \textbf{Implementation details.} 
We adopt Mask2Former~\cite{cheng2021masked} with ResNet-101~\cite{he2016deep} backbone as the proposal extractor and $\text{BERT}_{\text{base}}$~\cite{devlin2018bert} 
as the language encoder.
Following previous work~\cite{yu2018mattnet,luo2020multi,ding2021vision}, the Mask2Former is pre-trained on MS-COCO and we remove all images that appeared in val/test sets of RefCOCO/+/g. 
We adopt a $4$-layers transformer decoder for both indicated generation head and proposal selection head. 
For disambiguation supervision, we set $\tau$=$2.5$, $\alpha$=$4$, and sample $q_{1}$=$5$ expressions for Eq.(\ref{eq:reinforce}). For generation supervision, we generate $q_{2}$=$3$ expressions for each instance, filter instance with area$<$$32^{2}$, and set the pseudo-weight as $0.1$. The training details are in the Appendix.

\subsection{Results: Referring Expression Generation}

\begin{table}[htbp]
\centering
\footnotesize
\resizebox{8cm}{!}{
\begin{tabular}{l|ccccc}
\hline \toprule
 \multirow{2}{*}{Method} & \multicolumn{2}{c}{RefCOCO} & \multicolumn{2}{c}{RefCOCO+} & RefCOCOg \\
 & testA & testB & testA & testB & val \\ 
\hline \toprule
MMI~\cite{mao2016generation} & 66\% & 65\% & 34\% & 34\% & - \\
Attr~\cite{liu2017referring} & 78\% & 83\% & 41\% & 43\% & - \\
Attention~\cite{liu2020attribute} & 83\% & \underline{87\%} & 49\% & 46\% & - \\
PFOS~\cite{sun2022proposal} & 87\% & 84\% & 55\% & 53\% & 61\% \\ \hline
Ours (Box) & \underline{87\%} & 86\% & \underline{59\%} & \underline{58\%} & \underline{66\%} \\
Ours (Mask) & \textbf{89\%} & \textbf{87\%} & \textbf{61\%} &\textbf{ 61\%} & \textbf{67\%} \\
\bottomrule \hline
\end{tabular}
}
\vspace{-5pt}
\caption{Human evaluation for Referring Expression Generation (REG). `-' represents that the result is not provided.}
\vspace{-15pt}
\label{tab:human_eval}
\end{table}

\noindent \textbf{Quantitative Evaluation for REG.} Tab.~\ref{tab:reg_main_tab} reports the performance of REG tasks under automatic metrics.
We provide results using mask and box as input. The box results are obtained by taking the region inside the bounding box as a mask for Mask2Former training, which is a fair comparison to box-input methods.
Among all three datasets, our method outperforms previous work by a clear margin. Compare to the best competitors CoNAN~\cite{kim2020conan}, we show relative improvements of $3.3\%$ and $2.4\%$ on RefCOCO and RefCOCO+ under the main CIDEr metric. 
We find using the mask as input brings better results than using the box. We conjecture that the mask contains more precise object contour information, which helps to focus on the object foreground and avoid the interference of overlapping objects.

Besides automatic metrics, the unambiguity of the generated expression is crucial for REG.  Following previous works~\cite{mao2016generation,liu2017referring,liu2020attribute,sun2022proposal}, we conduct the human evaluation to evaluate the expression unambiguity: \textit{we randomly select $100$ generated expressions for each test split. For each expression, two human users are asked to click on the corresponding object in the image. The expression is considered unambiguous if both users click the ground-truth object.}
As shown in Tab.~\ref{tab:human_eval}, our generated expressions show superior unambiguity compared to previous work.
Since RefCOCO allows describing objects with simple location words (e.g., left and right), the previous method achieves a high accuracy of $87\%$ on both test splits, and our method achieves comparable performance.
RefCOCO+ forbids simple location words and relies more on object attributes, which is much more difficult. By fully exploring all objects in the scene, our model achieves the accuracy of $59\%$ and $58\%$. For RefCOCOg, our model also handles long descriptions well and achieves better accuracy of $66\%$ than PFOS.


\noindent \textbf{Qualitative Evaluation for REG.} 
Fig.~\ref{fig:vis} shows the qualitative results of our REG model.
In the first three rows, we show the generated expressions when training on different datasets. The target objects are in green contour. 
The generated expressions are concise in RefCOCO and RefCOCO+ since the reference expressions are short sentences.
Our model makes good use of location words on RefCOCO, and extracting all proposals first helps better generate descriptions like \textit{``second from the right''}.
On RefCOCO+, our model can also generate accurate descriptions of attributes and relationships. 
Since expressions in RefCOCOg are longer sentences, our model generates expressions with more details.
In the last row, we show how the proposed indicator module works. Given the image of $5$ detected buses whose contours are marked in different colors, by assigning the positive indicator to one bus and the negative indicator to others, our model generates accurate descriptions for each bus based on their locations.

\begin{table}[htbp]
\centering
\resizebox{8cm}{!}{
\begin{tabular}{c|cccc|cccc}
& \textbf{E2E} & \textbf{Joint} & \textbf{Ind.} & \textbf{D-Sup.} & Meteor & CIDEr &  H-Acc & Rb-Acc\\ \hline \toprule
(a) & &  &  &  & 0.226 & 0.791 & - & - \\
(b) &$\surd$ &  &  &  & 0.238 & 0.847 & - & - \\ \hline
(c) &$\surd$ & $\surd$ &  &  & 0.251 & 0.879 & 55\% & 73.5\%  \\
(d) &$\surd$ & $\surd$ & $\surd$ &  & 0.252 & 0.895 & 58\% & 75.6\% \\
(e) &$\surd$ & $\surd$ & $\surd$ & $\surd$ & 0.248 & 0.927 & 61\% & 79.1\%  \\ \bottomrule \hline
\end{tabular}
}
\vspace{-5pt}
\caption{Component ablations for REG on RefCOCO+ testB.}
\vspace{-15pt}
\label{tab:reg_componet}
\end{table}

\noindent \textbf{Component Ablations for REG.} As shown in Tab.~\ref{tab:reg_componet}, we investigate the effectiveness of each component on RefCOCO+ testB, which is the hardest test split. The baseline model (a) contains the proposal extractor but generates expressions based only on target object features.
End-to-end optimizing the proposal extractor for the REG task in (b) improves CIDEr to $0.847$, and joint training with the RES task in (c) further improves it to $0.879$. 
We propose a novel indicator module and disambiguation supervision to generate ambiguous expressions. As shown in (d) and (e), these two components effectively boost the CIDEr score to $0.895$ and $0.927$. 
We conduct the human evaluation for (c), (d), and (e), the accuracy (H-Acc) improvement is consistent with the automatic metrics.
Since human evaluation is subjective, we propose a model-based evaluation named Refer-back Accuracy (Rb-Acc), in which the expression $Y$ for $k$-th object is treated as correct only if $\operatorname*{arg\, max}_{i} S_{i}(Y)==k$, 
where $\mathcal{S}(Y)$ is matching scores provided by an additionally trained proposal selection head in Eq.(\ref{eq:matching_score}). As we can see, the indicator module and disambiguation supervision similarly improve the Rb-Acc from $73.5\%$ to $75.6\%$ and $79.1\%$.

\begin{table*}[htbp]
\centering
\footnotesize
\resizebox{15.1cm}{!}{
\begin{tabular}{l|c|c|ccc|ccc|cc}
\multicolumn{11}{l}{\textbf{\textit{Referring Expression Segmentation (RES):}}} \\
\hline \toprule
\multirow{2}{*}{Methods} & \multirow{2}{*}{Backbone} & \multirow{2}{*}{Pretrain} & \multicolumn{3}{c|}{RefCOCO} & \multicolumn{3}{c|}{RefCOCO+} & \multicolumn{2}{c}{RefCOCOg}\\
\cline{4-11}
& & & val & testA & testB & val & testA & testB & val & test \\ \hline \toprule
 MattNet~\cite{yu2018mattnet} & MRCN-101 & COCO & \cellcolor{mygray}56.51 & \cellcolor{mygray}62.37 & \cellcolor{mygray}51.70 & \cellcolor{mygray}46.67 & \cellcolor{mygray}52.39 & \cellcolor{mygray}40.08 & \cellcolor{mygray}47.64 & \cellcolor{mygray}48.61 \\
 CAC~\cite{chen2019referring} & MRCN-101 & COCO & \cellcolor{mygray}58.90 & \cellcolor{mygray}61.77 & \cellcolor{mygray}53.81 & - & - & -  & - & - \\
 NMTree~\cite{liu2019learning} & MRCN-101 & COCO & \cellcolor{mygray}56.59 & \cellcolor{mygray}63.02 & \cellcolor{mygray}52.06 & \cellcolor{mygray}47.40 & \cellcolor{mygray}53.01 & \cellcolor{mygray}41.56 & \cellcolor{mygray}46.59 & \cellcolor{mygray}47.88 \\
 ReSTR~\cite{kim2022restr} & ViT-B-16	& ImgNet-21K & \cellcolor{mygray}67.22 & \cellcolor{mygray}69.30 & \cellcolor{mygray}64.45 & \cellcolor{mygray}55.78 & \cellcolor{mygray}60.44 & \cellcolor{mygray}48.27 & - & - \\ \hline
 Ours & ResNet-101 & COCO & \cellcolor{mygray}\textbf{71.44} & \cellcolor{mygray}\textbf{73.80} & \cellcolor{mygray}\textbf{68.58} & \cellcolor{mygray}\textbf{61.59} & \cellcolor{mygray}\textbf{64.27} & \cellcolor{mygray}\textbf{56.08} & \cellcolor{mygray}\textbf{62.12} & \cellcolor{mygray}\textbf{62.81} \\ 
 \hline \toprule
 MCN~\cite{luo2020multi} & DarkNet-53 & COCO & 62.44 & 64.20 & 59.71 & 50.62 & 54.99 & 44.69 & 49.22 & 49.40 \\
 CGAN~\cite{luo2020cascade} & DarkNet-53 & COCO & 64.86 & 68.04 & 62.07 & 51.03 & 55.51 & 44.06 & 51.01 & 51.69 \\
 LTS~\cite{jing2021locate} & DarkNet-53 & COCO & 65.43 & 67.76 & 63.08 & 54.21 & 58.32 & 48.02 & 54.40 & 54.25 \\
 ISFP~\cite{liu2022instance} & DarkNet-53 & COCO & 65.19 & 68.45 & 62.73 & 52.70 & 56.77 & 46.39 & 52.67 & 53.00 \\
 VLT~\cite{ding2021vision} & DarkNet-53 & COCO & 65.65 & 68.29 & 62.73 & 55.50 & 59.20 & 49.36 & 52.99 & 56.65 \\
 CRIS~\cite{wang2021cris} & ResNet-101 & CLIP & 70.47 & 73.18 & 66.10 & 62.27 & 68.08 & 53.68 & 59.87 & 60.36 \\
 RefTR~\cite{li2021referring} & ResNet-101 & ImageNet & 70.56 & 73.49 & 66.57 & 61.08 & 64.69 & 52.73 & 58.73 & 58.51 \\
 \textit{RefTR$^{\dag}$}~\cite{li2021referring} & \textit{ResNet-101} & \textit{VG (4.29M)} & \textit{74.34} & \textit{76.77} & \textit{70.87} & \textit{66.75} & \textit{70.58} & \textit{59.40} & \textit{66.63} & \textit{67.39} \\
 \hline
 Ours& ResNet-101 & COCO & \textbf{75.57} & \textbf{77.17} & \textbf{72.73} & \textbf{67.80} & \textbf{70.65} & \textbf{62.09} & \textbf{68.08} & \textbf{68.57} \\
\hline \toprule 
\multicolumn{11}{l}{\textit{$^{*}$Referring Expression Comprehension (REC):}} \\ 
\hline \toprule
 TransVG~\cite{deng2021transvg} & ResNet-101 & COCO & 80.83 & 83.38 & 76.94 & 68.00 & 72.46 & 59.24 & 68.71 & 67.98 \\
 RefTR~\cite{li2021referring} & ResNet-101 & ImageNet & 82.23 & 85.59 & 76.57 & 71.58 & 75.96 & 62.16 & 69.41 & 69.40 \\ 
  \textit{RefTR$^{\dag}$}~\cite{li2021referring} & \textit{ResNet-101} & \textit{VG (4.29M)} & \textit{85.65} & \textit{88.73} & \textit{81.16} & \textit{77.55} & \textit{82.26} & \textit{68.99} & \textit{79.25} & \textit{80.01} \\
 \hline
 Ours & ResNet-101 & COCO & 83.74 & 87.32 & 81.27 & 72.84 & 78.04 & 65.80 & 73.61 & 73.09 \\ 
 \bottomrule \hline
\end{tabular}
}
\vspace{-5pt}
\caption{Comparison with state-of-the-art methods for Referring Expression Segmentation (RES). `-' represents that the result is not provided. Results in \colorbox{mygray}{gray shadow} are measured by oIoU metric and others are measured by mIoU metric.}
\vspace{-15pt}
\label{tab:res_main}
\end{table*}

\begin{table}[htbp]
\centering
\resizebox{7.4cm}{!}{
\begin{tabular}{l|l|ccc}
& \textbf{Reward} & Meteor & CIDEr & Rb-Acc \\ \hline \toprule
(a) & Unambiguity & 0.241$\downarrow$ & 0.844$\downarrow$ & 80.5\% \\
(b) & Unambiguity + CIDEr & 0.248 & 0.927 & 79.1\% \\ \hline
(c) & Sigmoid & 0.246 & 0.914 & 74.5\%$\downarrow$ \\
(d) & Softmax & 0.250 & 0.926 & 78.0\% \\
(e) & Softmax + $\tau$ & 0.248 & 0.927 & 79.1\% \\ \bottomrule \hline
\end{tabular}
}
\vspace{-5pt}
\caption{Ablation of different rewards in disambiguation supervision on RefCOCO+ testB.}
\vspace{-20pt}
\label{tab:reward}
\end{table}

\noindent \textbf{Effectiveness of Disambiguation Supervision.} We utilize reinforcement learning to improve REG performance, in which the unambiguity reward is provided by our proposal selection head. In Tab.~\ref{tab:reward}, we compare the result with different rewards on RefCOCO+ testB. 
As shown in (a), utilizing only the unambiguity reward improves the Rb-Acc but decreases Meteor and CIDEr scores, while adding the CIDEr reward in (b) avoids such performance drop.
As we proposed in Eq.(\ref{eq:reward}), the Softmax-based unambiguity reward compares the matching scores across all objects, which is more robust. Compare with the Sigmoid-based reward (adopted by JointSLR~\cite{yu2017joint}) in (c), we achieve better results under both automatic metrics and Rb-Acc in (d). And including $\tau$ to smooth the matching score in (e) further improves the generated expressions.

\subsection{Results: Referring Expression Segmentation}

\noindent \textbf{Quantitative Evaluation for RES.} Tab.~\ref{tab:res_main} shows the quantitative results for RES.
For a fair comparison, most of the methods involved in the comparison are pre-trained on MS-COCO~\cite{lin2014microsoft}, and the computational amounts of the backbones (ResNet-101 and DarkNet-53) are similar.
Since previous top-down methods report their results under oIoU but recent bottom-up ones report under mIoU, we report results under both metrics.
Compare with the best competitor ReSTR~\cite{kim2022restr} under the oIoU metric, we achieve an average $5.05\%$ improvement on all test splits. And compare with RefTR~\cite{li2021referring} and CRIS~\cite{wang2021cris} under the mIoU metric, we also achieve an average $5.97\%$ improvement.
We achieve greater performance boosts on harder dataset RefCOCO+ than on RefCOCO, which reflects a better vision-language understanding of our model. 
Besides, we also compare with models under stronger training settings. As shown in Tab.~\ref{tab:res_main}, since RefTR jointly trains the RES and REC tasks, it leverages Visual Genome (VG) for pretraining, which contains about $4.29M$ box-expression pairs. Our model still outperforms it without external annotations. 
And under the setting using Swin-Transformer~\cite{liu2021swin} backbone, we also outperform LAVT~\cite{yang2021lavt}, the results are in Appendix.
The superior performance comes from the stronger pipeline, joint training with REG, and the proposed mutual supervision. 
The visualizations of our RES model are shown in Appendix.

We also report the results of Referring Expression Comprehension (REC) task in Tab.~\ref{tab:res_main}, by simply outputting the bounding box of our predicted mask. Under the REC metric that calculates the ratio of box-IoU$>$0.5, our model is also comparable with RefTR.

\begin{table}[htbp]
\centering
\resizebox{8cm}{!}{
\begin{tabular}{c|ccc|ccc|c}
 & \textbf{E2E} & \textbf{Joint} & \textbf{G-Sup.} & Ref. & Ref+. & Refg. & $\Delta$ \\ \hline \toprule 
(a) &  &  &  & 68.98 & 59.23 & 61.47 & -\\
(b) & $\surd$ &  &  & 74.21 & 64.99 & 66.42 & - \\ \hline
(c) & $\surd$ & $\surd$ &  & 74.83 & 66.21 & 66.91 & +0.78\\
(d) & $\surd$ &  & $\surd$ & 74.91 &	67.24 &	67.85 & +1.46\\ \hline
(e) & $\surd$ & $\surd$ & $\surd$ & 75.57 & 67.80 & 68.08 & +1.94 \\ \bottomrule \hline
\end{tabular}
}
\vspace{-5pt}
\caption{Component ablations for RES on validation sets of all three datasets. $\Delta$ calculates the average improvement over (b).}
\vspace{-15pt}
\label{tab:res_abl}
\end{table}

\noindent \textbf{Component Ablations for RES.}  
Tab.~\ref{tab:res_abl} shows the component ablations on validation sets of all three datasets. 
The baseline model in (a) is similar to MattNet~\cite{yu2018mattnet},  which first generates all object proposals and selects the one that best matches the expression. We use an advanced segmentation network Mask2Former~\cite{cheng2021masked} instead of the Mask-RCNN~\cite{he2017mask}, which leads to better results.
We then show the importance of end-to-end optimization. Like MattNet, object proposal features in (a) are only supervised by instance segmentation loss, which has a gap with the RES task. In (b), end-to-end optimization with RES loss brings over $5\%$ performance boost.
In (c) and (d), we show that the joint training with the REG task and our proposed generation supervision can bring average improvements of $0.78\%$ and $1.46\%$ over (b), respectively. 
The generation supervision performs better on more difficult datasets, boosting $2.25\%$ mIoU on RefCOCO+. 
We combine all components in (e), which shows further improvement.

\begin{figure}[htbp]
  \centering
  \includegraphics[width=0.37\textwidth]{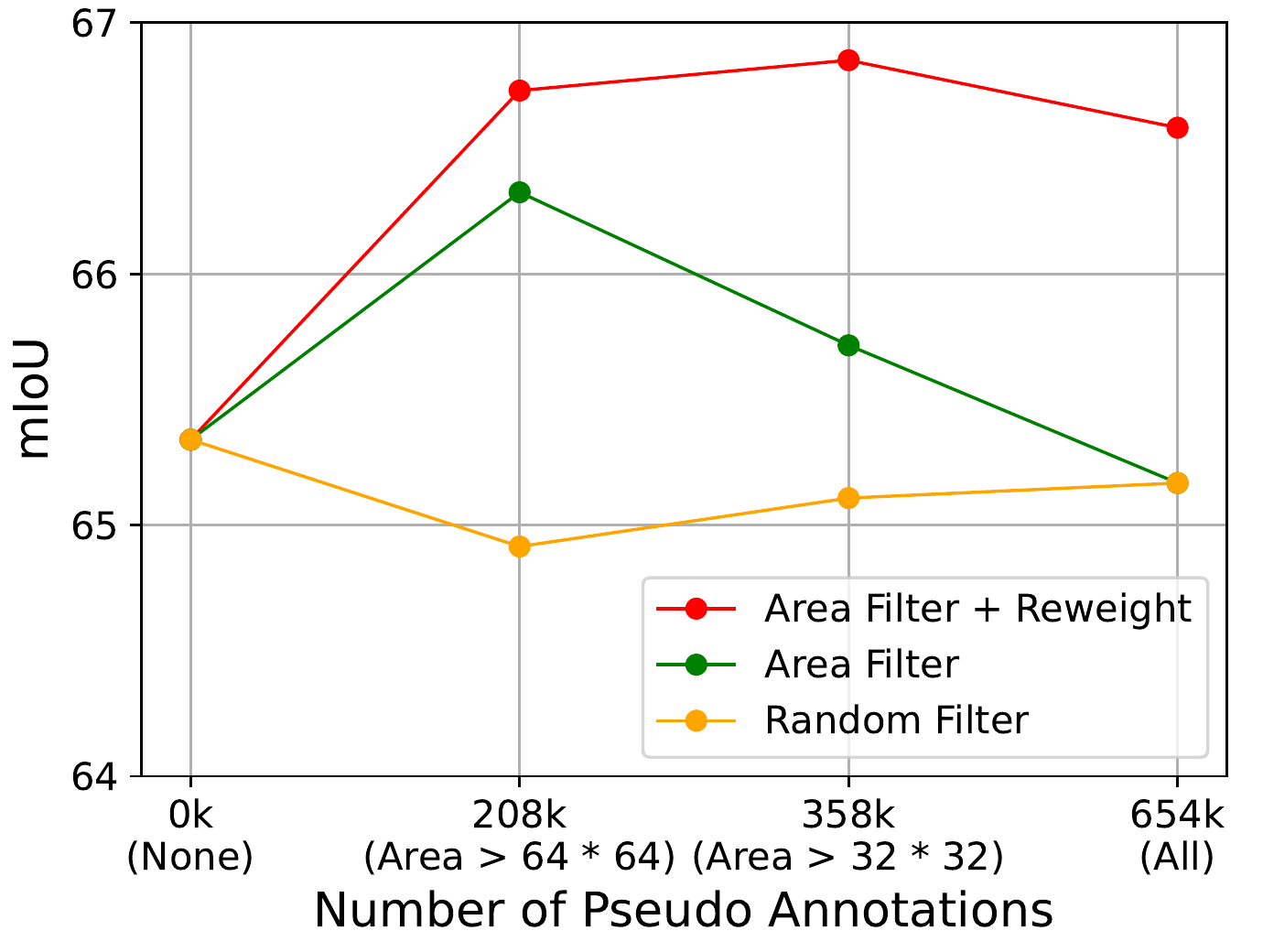}
  \vspace{-10pt}
  \caption{Ablation studies for data filtering and data reweighting on RefCOCO+, we report the average results on three test splits.}
  \vspace{-15pt}
  \label{fig:pse_abl}
\end{figure}

\begin{table}[htbp]
\centering
\resizebox{7.4cm}{!}{
\begin{tabular}{c|ccc}
\textbf{Pseudo-weight} & Ref+. val & Ref+. testA & Ref+. testB \\ \hline \toprule
0.05 & 67.69 & 70.63 & 61.71 \\
0.1 & \textbf{67.80} & \textbf{70.65} & \textbf{62.09} \\
0.5 & 67.37 & 70.42 & 61.62 \\
1.0 & 66.58 & 69.72 & 60.83 \\ \bottomrule \hline
\end{tabular}}
\vspace{-8pt}
\caption{Ablation studies of different pseudo-weights in generation supervision on RefCOCO+.}
\vspace{-10pt}
\label{tab:pse_weight}
\end{table}

\noindent \textbf{Effectiveness of Generation Supervision.} 
The proposed generation supervision improves RES performance by generating expressions on unannotated COCO instances, in which the area-based data filtering and data reweighting strategies are keys to work. 
As shown in Fig.~\ref{fig:pse_abl}, we analyze the effectiveness of these two strategies on the RefCOCO+ dataset and report the average mIoU across all three test splits. The baseline model was trained without pseudo-annotations (None). Firstly for data filtering, we compare three filter rules, including not filtering data (All), filtering out instances with area less than $32^2$ (Area$>$$32^2$), and filtering out instances with area less than $64^2$ (Area$>$$64^2$). These three rules preserve $654k$, $358k$, and $208k$ pseudo-annotations respectively.
We introduce random filtering for comparison. As we can see, after joint training with $208k$ pseudo-annotations via random filtering, the performance does not improve but declines. This is because the noise in pseudo-data hurts the training. The performance gradually improves as we use more pseudo-annotations, but still lags behind the baseline.
Instead, with our area-based filtering, performance improves as we gradually filter out annotations with areas smaller than $32^2$ and $64^2$. This is because the expressions generated with larger objects are more accurate, and also conform to the distribution of RES tasks.
Above the data filtering, the data reweighting strategy can further suppress the noise, in which we reduce the loss weight of pseudo-annotations to $0.1$. Combining these two strategies, we find Area$>$$32^2$ achieves the best performance, which is a balance between the data amount and data noise.
As shown in Tab.~\ref{tab:pse_weight}, we also test results with different loss weights of the pseudo annotations, the pseudo weight of $0.1$ achieves the best performance.

\begin{table}[htbp]
\centering
\resizebox{7.4cm}{!}{
\begin{tabular}{c|c|ccc}
\textbf{Resolution} & mIoU & Pr@0.5 & Pr@0.7 & Pr@0.9 \\ \hline \toprule
  $416 \times 416$ & 73.53 & 82.09 & 74.92 & 33.36 \\
  $640 \times 640$ & 75.27 & 83.33 & 76.89 & 39.82 \\
  $1024 \times 1024$ & 75.57 & 83.18 & 77.94 & 41.12 \\ \bottomrule \hline
\end{tabular}
}
\vspace{-8pt}
\caption{RES results under different resolutions on RefCOCO val.}
\vspace{-10pt}
\label{tab:resoltuion}
\end{table}

\noindent \textbf{Performance with Different Resolutions.} 
We adopt a resolution of $1024$ in the main results following previous top-down methods~\cite{yu2018mattnet,chen2019referring}.
In Tab.~\ref{tab:resoltuion}, we test performance under different resolutions by resizing the long side to $416/640$ and padding the short side, which are common settings for bottom-up methods~\cite{wang2021cris,li2021referring}. 
On RefCOCO val, resizing to $640$ leads to only a tiny performance drop to $75.27$, still superior to the best competitor RefTR of $70.56$. 
The much smaller resolution of $416$ caused Pr@0.9 to drop from $41.12$ to $33.36$, but the mIoU still achieves $73.53$.

\section{Conclusion and Future Work}
\label{sec:conclusion}

In this paper, we present a unified mutual supervision framework that jointly learns the RES and REG tasks, including a proposal selection head for RES and an indicated generation head for REG. Our framework learns under mutual supervision, \textit{i.e.}, disambiguation supervision, and generation supervision. Such unified mutual supervision effectively improves two tasks by solving their bottleneck problems.
We conduct detailed experiments on RefCOCO, RefCOCO+, and RefCOCOg. Our approach outperforms existing methods on both REG and RES tasks under fair comparison, and extensive ablation studies demonstrate the effectiveness of our architecture and the proposed mutual supervision.
In our future work, we will explore how to merge the current multi-step mutual supervision paradigm into one step and apply expression generation on larger scale data.


\clearpage

{\small
\bibliographystyle{ieee_fullname}
\bibliography{main}
}
\clearpage

\appendix

\section{Introduction of Datasets}
We conduct extensive experiments on three
benchmark datasets for both REG and RES tasks, including RefCOCO~\cite{yu2016modeling}, RefCOCO+~\cite{yu2016modeling}, and RefCOCOg~\cite{mao2016generation}, which are all collected from MS-COCO~\cite{lin2014microsoft}. The RefCOCO dataset contains 19,994 images with 142,209 referring expressions for 50,000 objects while the RefCOCO+ dataset contains 19,992 images with 141,564 expressions for 49,856 objects. RefCOCO+ has no location words and is more challenging than RefCOCO. RefCOCOg consists of 26,711 images with 104,560 referring expressions for 54,822 objects. The expressions are of an average length of 8.4 words which is much longer than other datasets. RefCOCOg has two types of data partitions, i.e., Google partition~\cite{mao2016generation} and UNC partition~\cite{nagaraja2016modeling}. For the REG task, we adopt Google partition to compare with previous work. For the RES task, we adopt the UNC partition and benchmark our model with other methods using the same partitions for fair comparisons.


\section{Training Details}
We implement our code on Detectron2~\cite{wu2019detectron2}. We adopt Mask2Former~\cite{cheng2021masked} with ResNet-101 backbone as the proposal extractor and $\text{BERT}_{\text{base}}$~\cite{devlin2018bert} (implemented by HuggingFace~\cite{wolf2020transformers}) as the language encoder.
We adopt AdamW~\cite{loshchilov2017decoupled}  with a weight decay of $0.05$ as the optimizer.
We train the model in three steps. In step 1, we train the entire network with a learning rate of $5e\text{-}4$ for $90$k iterations with $8$ images in a batch. The learning rate is multiplied by $0.1$ at $60$k and $80$k iterations. In step 2, we freeze the proposal extractor and the proposal selection head and train the indicated generation head with a learning rate of $1e\text{-}6$ for $20$k iterations with $4$ images in a batch. In step 3, we retrain the entire network on the dataset combining manually-annotated expressions and generated expressions, with the same training configuration of step 1.


\section{RES Performance using Swin-Transformer}

\begin{table*}[htbp]
\centering
\resizebox{13.9cm}{!}{
\begin{tabular}{l|c|ccc|ccc|cc}
\hline \toprule
\multirow{2}{*}{Methods} & \multirow{2}{*}{Backbone} & \multicolumn{3}{c|}{RefCOCO} & \multicolumn{3}{c|}{RefCOCO+} & \multicolumn{2}{c}{RefCOCOg}\\
& & val & testA & testB & val & testA & testB & val & test \\ \hline
LAVT~\cite{yang2021lavt} & Swin-B & 74.46 & 76.89 & 70.94 & 65.81 & 70.97 & 59.23 & 63.34 & 63.62 \\
Ours & Swin-B & 77.51 & 78.87 & 74.68 & 68.96 & 73.62 & 62.56 & 68.92 & 68.95 \\
 \bottomrule \hline
\end{tabular}
}
\caption{Comparison with LAVT for Referring Expression Segmentation (RES), under mIoU metric.}
\vspace{10pt}
\label{tab:res_swin}
\end{table*}

The stronger visual backbone can bring a significant improvement to the RES task. We provide RES results using Swin-Transformer~\cite{liu2021swin} as the visual backbone in Tab.~\ref{tab:res_swin}, both the Swin-Transformer backbone of LAVT and ours are pre-trained on ImageNet-22K.
Using Swin-B as the backbone, LAVT~\cite{yang2021lavt} achieves previous state-of-the-art performance. Our RES network performs better than LAVT. Using ResNet101 as the visual backbone, our model is already outperforms with LAVT on RefCOCO and RefCOCO+, and RefCOCOg. Using the Swin-B as the visual backbone, our model further outperforms LAVT by a large margin.

\section{Discussion about Assets}
We implement the code based on detectron2~\cite{wu2019detectron2}, which is released under the Apache 2.0 license. We run experiments on three benchmark datasets RefCOCO~\cite{yu2016modeling}, RefCOCO+~\cite{yu2016modeling}, and RefCOCOg~\cite{mao2016generation}. All of them are created based on MS-COCO~\cite{lin2014microsoft}. The annotations of MS-COCO belong to the COCO Consortium and are licensed under a Creative Commons Attribution 4.0 License. The use of images in MS-COCO must abide by the Flickr Terms of Use. The users of the images accept full responsibility for the use of the dataset, including but not limited to the use of any copies of copyrighted images that they may create from the dataset. The MS-COCO dataset contains a category named person, and images in this category may contain personally identifiable information.

\section{Qualitative Evaluation for RES}

\begin{figure*}[htbp]
  \centering
  \includegraphics[width=0.9\textwidth]{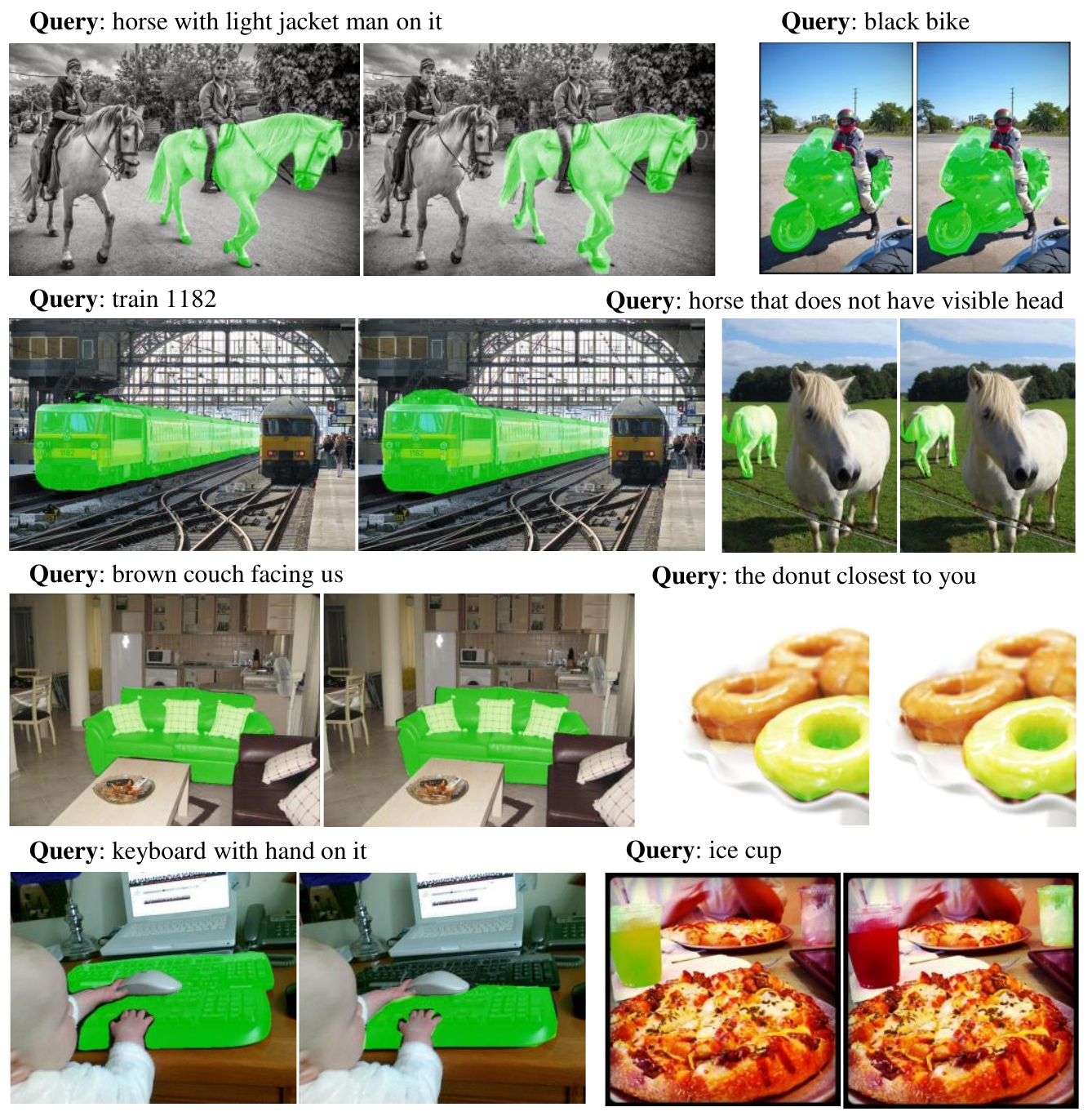}
  \caption{Qualitative Evaluation for Referring Expression Segmentation (RES). For each query, the model prediction and the ground truth are given in order. The segmentation masks are in green color.}
  \label{fig:sup_vis2}
\end{figure*}

We show the qualitative evaluation of our RES model in Fig.~\ref{fig:sup_vis2}. The visual backbone is ResNet-101 and the language encoder is BERT. For each query, we show the model prediction and ground-truth mask from left to right. Since the top-down pipeline decouples the segmentation prediction and proposal selection, our model shows high-quality segmentation results, and also a good understanding between vision and language. In the last row of Fig.~\ref{fig:sup_vis2}, we show two failure cases of our RES model. The reason for the failure cases comes from the proposal extractor providing inaccurate segmentation and the proposal selection head selecting the wrong proposal with the given query.





\end{document}